\documentclass[11pt]{article}

\usepackage[preprint]{acl}

\usepackage{times}
\usepackage{latexsym}

\usepackage[T1]{fontenc}

\usepackage[utf8]{inputenc}

\usepackage{microtype}

\usepackage{inconsolata}

\usepackage{graphicx}

\usepackage{booktabs}

\usepackage[colorinlistoftodos,prependcaption,textsize=tiny,textwidth=14mm]{todonotes}
\makeatletter
\define@key{todonotes}{CY}[]{%
    \setkeys{todonotes}{author=CY,color=cyan!20}}%
\makeatother   
\makeatletter
\define@key{todonotes}{AJ}[]{%
    \setkeys{todonotes}{author=AJ,color=purple!20}}%
\makeatother   
\makeatletter
\define@key{todonotes}{SM}[]{%
    \setkeys{todonotes}{author=SM,color=green!20}}%
\makeatother

\newcommand{\dname}{\texttt{CREATE} }

\title{\texttt{CREATE}: Creativity Evaluation of VLMs using the Visual Remote Associates Test}

\author{Ziqing Yang$^{*1}$, Anubhav Jangra$^{*1}$, Smaranda Muresan$^{2}$\\
$^1$Columbia University, USA\\
$^2$Barnard College, Columbia University, USA\\
\texttt{zy2644@columbia.edu}, \texttt{anubhav@cs.columbia.edu}, \texttt{smuresan@barnard.edu}
}

\begin{document}
\maketitle
\begingroup
\def\thefootnote{*}\footnotetext{ denotes equal contribution}
\endgroup
\begin{abstract}
Vision language models (VLMs) are used in creative tasks such as visual storytelling, design ideation, art 
interpretation, and advertising. Yet we lack principled ways to measure the lateral reasoning of these models compared to human potential. 
We introduce \texttt{CREATE}, a semi-automatically constructed benchmark containing 155 visual Remote Associates Test problems. These problems require a solver to find an underlying connection among three seemingly dissimilar images. We conduct a human study (N=25) to validate and ground the quality of our benchmark, and evaluate 11 VLMs. Our findings show that most VLMs underperform humans on this task. 
Our experiments indicate that the associative ability of VLMs depends on the cue concepts provided in the text and does not universally extend to visual puzzles.\footnote{Implementation and data can be found at \url{https://github.com/CeYang214/CREATEivity-benchmark}.}
\end{abstract}

\section{Introduction}

Puzzles have served as instruments for measuring human creativity, requiring solvers to make connections that are not obviously related~\citep{guilford1950creativity, maier1931reasoning}. A key aspect of this creative puzzle solving is `insight', the sudden restructuring of a problem that brings a previously hidden solution into view, which often tends to arrive abruptly in short puzzles~\cite{kounios2014cognitive, metcalfe1987intuition}. The Remote Associates Test (RAT) is a widely used framework for studying insight and human creativity ~\citep{Mednick1962TheAB}, where a solver is given three \textit{cue words} (\textit{e.g.,} \textit{sandwich}, \textit{golf}, and \textit{house}) and must find a fourth word that links them all (\textit{e.g.,} (\textit{club})). \citet{oltecteanu2020visual} has extended these problems into the visual domain by replacing the cue words with cue images. While these visual RATs are an attractive candidate for evaluating creativity VLMs, existing RAT problems consist of fixed, manually curated sets reused for over 5 decades.

\begin{figure}[t]
    \centering
    \includegraphics[width=\columnwidth]{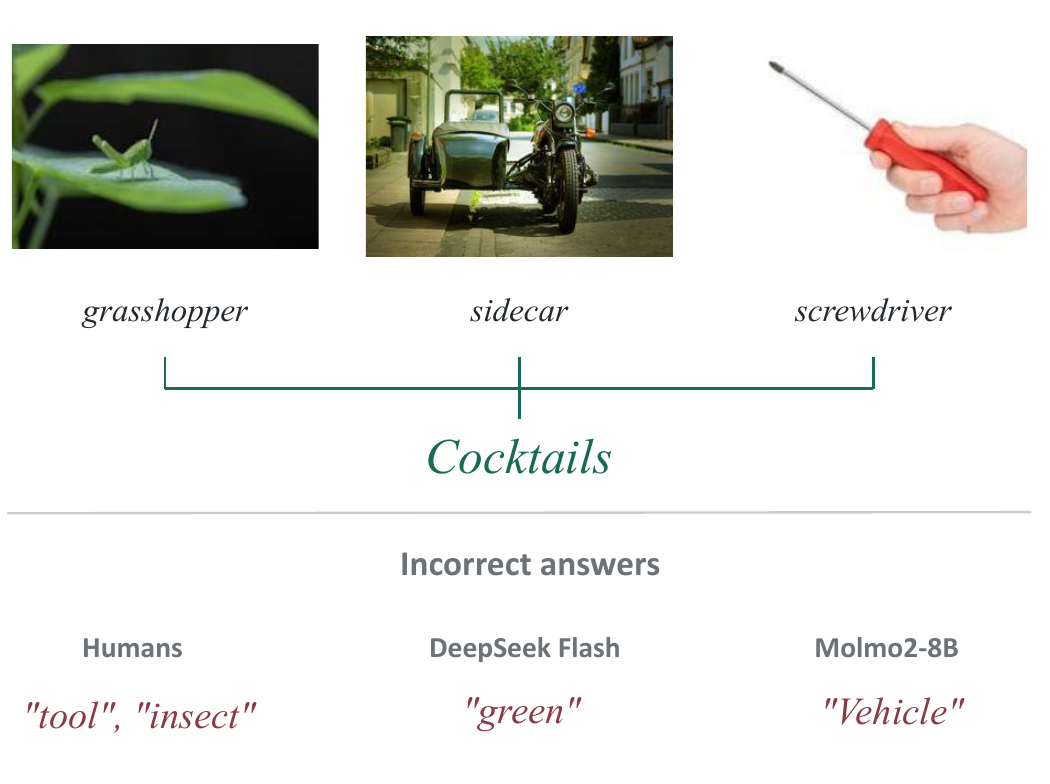}
    \caption{An example from the \texttt{CREATE} benchmark. The three \textit{cue images} show a ``grasshopper", a ``sidecar", and a ``screwdriver", connected by the \textit{answer word} ``cocktail'' (all three are types of cocktails).}
    \label{fig:puzzle-example}
\end{figure}

To bridge this gap, we propose a human-AI collaboration framework to construct the \dname benchmark (piCture REmote Associates TEst) comprising of 155 puzzles (see Figure~\ref{fig:puzzle-example} for an example), and position visual RAT problems as an evaluation framework for creativity in VLMs. We ask two research questions, \textbf{RQ1}: How do humans and VLMs differ in solving visual RAT problems? \textbf{RQ2}: Do VLMs rely on language as an intermediate step when solving visual RAT problems? For a subset of 80 problems, we conduct a human study with 25 participants, each of whom attempted 16 problems. Complementing this, we evaluat seven open-weight and four proprietary VLMs.

Our human study shows that, on average, humans solve 22.8\% of problems, 
with 70\% of the problems solved by at least one of the five participants. Only one VLM outperforms humans on this task, namely, \texttt{Claude-Opus-5.5}, which solves 32.5\% of the puzzles in a zero-shot setting. It solves 4 puzzles that no 
human participant solves, but fails on 34 puzzles that at least one human participant solves. When VLMs answer incorrectly, they are more than twice as likely as humans to describe what the images literally depict. 
Presenting the same puzzles as text-only RAT problems raised every model's accuracy by at least 30 points, with the best model reaching 90\%. The models could therefore find the association once the cue words were given, but could not reliably recover those words from the images, which points to visual abstraction as their principal limitation.
This work makes three key contributions. \textbf{First}, we provide the \dname benchmark, comprising 155 visual RAT problems. \textbf{Second}, we conduct a human study to validate and ground each instance in human responses. \textbf{Third}, we evaluate 11 VLMs and show they 
underperform compared to human performance. Our analysis reveals that their errors 
are likely due to failed visual abstraction rather than from the association reasoning itself.

\section{Related Work}

Prior work has developed textual and visual puzzle benchmarks to test different reasoning capabilities of LLMs and VLMs. Wordplay-style text puzzles include the work by ~\citet{samdarshi-etal-2024-connecting}, who built a puzzle benchmark inspired by The New York Times' Connection\footnote{https://www.nytimes.com/games/connections} to evaluate models on the task of organizing 16 words into four groups of four words with shared associations within each group.~\citet{jiang-etal-2023-brainteaser} developed \texttt{BRAINTEASER}, a multiple choice question answering dataset designed to test models' ability to exhibit lateral thinking defying commonsense reasoning. \citet{chia-etal-2024-puzzlevqa} developed \texttt{PuzzleVQA} to assess the model's visual perception and pattern completion, asking the model to complete missing patterns in colors, numbers, sizes, or shapes. \citet{ghosal2024languagemodelspuzzleprodigies} presented \texttt{AlgoPuzzleVQA}, evaluating VLMs on algorithmic puzzles requiring visual, language, and algorithmic reasoning capabilities.~\citet{xu-etal-2026-see} developed \texttt{MultiPun}, a benchmark comprising multi-modal puns, and found that VLMs struggle to distinguish genuine puns from distractors. \citet{lee-etal-2025-puzzled} and \citet{10.1609/aaai.v39i4.32464} similarly evaluated models on rebus puzzles that require recovering words or phrases from pictures, symbols, and spatial arrangements. 

\citet{wu2025bongardopenworldfewshotreasoningfreeform} released \texttt{Bongard-OpenWorld}, extending classical Bongard Problems~\citep{bongard1970pattern} to open-world images, where the task requires a solver to segregate 12 images into two sets of six based on a concept that sets them apart. Similarly,~\citet{raganato-etal-2023-semeval} presented the visual word sense disambiguation task, where the goal was to identify an image that corresponds to the intended meaning of a given ambiguous word out of ten candidate images. \citet{pmlr-vR16-aditya18a}'s work remains closest to ours, with their proposed \texttt{image-riddles} dataset comprising three images and the task of finding the common connection among them. However, we found that a significant portion of the riddles in the benchmark did not require cross-image associations. We found that captioning a single image with \texttt{Qwen3.5-4B} produced the exact answer for 42.9\% of riddles, and for 63.0\% once synonyms are accepted. 
\section{\texttt{CREATE}-ing the benchmark}\label{sec:dataset}

We propose a human-AI collaboration framework to build our benchmark. First, we used \texttt{Gemini-3.8-Flash} to construct text-only RAT puzzles (i.e., 3 cue words and an answer word that connects them). Specifically,  
we start with 532 seed domains spanning diverse areas such as food, space, and entertainment. For each domain, we prompted the model to propose several connecting \textit{answer words} and, for each answer word, a cluster of polysemous \textit{cue words} that all bear the same semantic relation to it (\textit{e.g.,} \textit{is a} or \textit{part of}, \textit{symbol of}). Each cue word had to  
have two distinct word senses: a domain sense that links it to the answer word, and an alternate sense that can be visually depicted as an object or a scene. For example, for the answer word \textit{cocktail}, the cue words included \textit{screwdriver} and \textit{grasshopper}, whose alternate senses are a tool and an insect, respectively. 


The model generated 3,340 cue-word candidates. In order to ensure that the cue words can be represented as adequate cue images in visual RAT puzzles, we apply a filtering step to account for high word familiarity, high imageability, and high distance between the two senses of the word. 
Familiarity of candidate cue word was computed as the frequency of unigram and bigram constituents over the English subset of the spoken language corpus OpenSubtitles~\cite{tiedemann2026opensubtitles2024} and Wikipedia dump 2026\footnote{\url{https://dumps.wikimedia.org/enwiki/}}. If unavailable in either corpus, we used the wordfreq Python package\footnote{\url{https://pypi.org/project/wordfreq/}} to assign a familiarity score. We fine-tuned a \texttt{BERT} model~\cite{devlin2019bert} on the MRC Psycholinguistic Database~\cite{coltheart1981mrc} to predict imageability of the cue words (Appendix~\ref{appx:imageability}). We used the \texttt{all-MiniLM-L6-v2} embeddings~\cite{wang2020minilm} to compute the cosine similarity between the cue word's alternate meaning and the candidate answer (\textit{e.g.,} ``screwdriver - tool'' and ``screwdriver - cocktail''). We manually annotated 152 candidate cue words and then trained a logistic regression model to learn the boundaries for all filters jointly. Post-filtering, we iteratively regenerated candidate cue words until each candidate puzzle had at least 3 acceptable candidates or reached 3 repair steps.

After filtering, we obtained 729 text-only RAT puzzles that were manually extended with images from the free-to-use photo collection platforms Unsplash and Pexels\footnote{\url{https://unsplash.com/} and \url{https://pexels.com/}}. We limited the puzzles in the final benchmark to three images, selecting the best-ranked candidate cue words when more than three were available using the logistic regression model. After manual image selection and verification, the final benchmark contains 155 puzzles spanning \textit{is-a}, \textit{part-of}, \textit{used-for}, and \textit{symbol-of} relations between the cue words and the connecting answer.

\section{Experimental Settings}

\noindent\textbf{Human Study.} We conducted a study with 25 participants to validate the quality and difficulty of our benchmark. The study was conducted on a subset of 80 puzzles, with each puzzle attempted by 5 participants. They were allowed up to three attempts to solve the puzzle, after which they were asked two questions about the association of the image and the cue word, and the cue word and the answer word. If they successfully solved the puzzle, an additional question was asked to rate the image's helpfulness in solving the puzzle (see Appendix~\ref{appx:ui} for the user interface). The post-quiz survey concluded with open-ended questions about their puzzle-solving strategies and study experiences.

\noindent\textbf{VLM evaluation.} To evaluate the creative reasoning of vision language models, we evaluated 7 open-weight models \texttt{Molmo2-8B} \citep{clark2026molmo2openweightsdata}, \texttt{DeepSeek-V4.1-Flash} \citep{deepseekai2026deepseekv41flashpushinglimitskv}, \texttt{Qwen3.8-27B}, \texttt{Qwen3.5-9B}, \texttt{Gemma-4-12B-it}, \texttt{InternVL3.5-8B} \citep{wang2025internvl35advancingopensourcemultimodal}, \texttt{Llama-4-Scout-17B (w4a16)} and 4 proprietary models \texttt{GPT-5.6 Luna}, \texttt{GPT-5.6 Sol}, \texttt{Claude Haiku 4.5}, and \texttt{Claude Opus 5.5}. We evaluated all models in five settings: \textit{zero-shot} prompt directly asks to give the shared concept, \textit{detailed} prompt provides details about the Remote Associates Task of finding an alternate meaning corresponding to the cue words, and \textit{stepwise} prompt asks the model to first identify the words for each image to convert it into a RAT problem. We also investigate the \textit{text-only} setting, where the VLMs are given directly the cue-words rather than cue-images. 
We use \texttt{Gemini-3.5-Flash-Lite} to assess answers in the human study and the generated VLM responses. Both generation and evaluation prompts are provided in Appendix~\ref{appx:prompt}. 


\section{Results and Discussion}\label{sec:results}

\begin{table}[t]
\centering
\small
\caption{Performance of humans and VLMs on the subset of 80 problems. VLMs were evaluated on five settings: \textit{zero-shot} (\textit{ZS}), \textit{detailed} (\textit{Det.}), \textit{stepwise} (\textit{SW}), \textit{two-shot} (\textit{2-S}) and \textit{text-only} (\textit{Txt.}).} \label{tab:task2-80}
\resizebox{\linewidth}{!}{%
\begin{tabular}{p{3cm}|ccccc}
\toprule
Model & \textit{ZS} & \textit{Det.} & \textit{SW} & \textit{2-S} & \textit{Txt.}\\
\midrule
Human (avg.) & 22.8 & -- & -- & -- & -- \\ 
Human ($\geq1$ correct) & 70.0 & -- & -- & -- & -- \\
\hline
\texttt{DeepSeek-Flash} & 5.0 & 13.8 & 5.0 & 15.0 & 63.8 \\
\texttt{Molmo2-8B} & 7.5 & 1.3 & 3.8 & 0.0 & 40.0 \\
\texttt{Qwen3.5-9B} & 0.0 & 1.3 & 1.3 & 5.0 & 36.3 \\
\texttt{Qwen3.8-27B} & 8.8 & 6.3 & 2.5 & 12.5 & 51.3 \\
\texttt{Gemma-4-12B-it} & 12.5 & 5.0 & 3.8 & 5.0 & 56.3 \\
\texttt{InternVL3.5-8B} & 1.3 &3.8 & 0.0 & 0.0 & 31.3 \\
\texttt{Llama-4-Scout-17B} & 1.3 & 2.5 & 1.3  & 0.0 & 43.8 \\\hline
\texttt{GPT-5.6-Luna} & 6.3 & 5.0 & 5.0 & 15.0 & 61.3 \\
\texttt{GPT-5.6-Sol} & 7.5 & 23.8 & 28.8 & 16.3 & 73.8 \\
\texttt{Claude-Haiku-4.5} & 5.0 & 2.5 & 0.0 & 26.3 & 60.0 \\
\texttt{Claude-Opus-5.5} & \textbf{32.5} & \textbf{52.5} & \textbf{43.8} & \textbf{45.0} & \textbf{90.0} \\
\bottomrule
\end{tabular}
}
\end{table}

To allow us to compare VLMs with human performance, we report 
the main results for the 80-puzzle subset in this section, with details on the full benchmark 
in Appendix~\ref{appx:results}.

\paragraph{RQ1: How do humans and VLMs differ?}
The participants, on average, correctly solved 16.5\% of the puzzles on the first attempt and 22.8\% by the third attempt, spending 83.6 seconds per puzzle. However, performance varied across participants: 4 did not solve any problems, 5 solved one, 4 solved two, and five participants accounted for nearly half of the correct solutions (49\%), with the best solving 12 of 16 puzzles. Participants solved 22.8\% puzzles on average, with 70\% puzzles being solved by at least one participant. Out of 80 puzzles, 
only 4 puzzles were solved by 3 out of 5 humans, while only one puzzle (cue words: \textit{bed}, \textit{bank}, \textit{mouth}, connecting word \textit{river}) was answered by all five participants. The study indicates that the puzzles are challenging yet solvable. They further indicate that solving visual RAT problems is not a universal ability, as correct solutions were concentrated among a small group of strong solvers, while most participants solved few or none of them.


We provide the human and VLM performance on our \dname benchmark in Table~\ref{tab:task2-80}. In the zero-shot setting, only one VLM outperformed the average human performance, with \texttt{Claude-Opus-5.5} solving 32.5\% of the puzzles. All other VLMs struggled to solve the puzzles, the next best model achieving 12.5\% accuracy. Of the 56 puzzles solved by at least one human participant, 22 were solved by \texttt{Claude-Opus-5.5}. 
There were four instances where \texttt{Claude-Opus-5.5} solved the puzzle when none of the participants could solve them. 

Prompting strategies had uneven effects relative to the zero-shot setting, with \textit{stepwise} prompting having a significant impact on \texttt{Claude-Opus-5.5} (+11.3 points), and \textit{detailed} prompt improving the performance of six models (\texttt{Claude-Opus-5.5}: +20.0, \texttt{GPT-5.6-Sol}: +16.3, \texttt{DeepSeek-Flash}: +8.8). The \textit{two-shot} setting improved the stronger reasoning model's performance, while reducing the three open-weight models to zero. In contrast, the text-only setting improved every model by 30.0 to 66.3 points, with \texttt{Claude-Opus-5.5} reaching 90.0\%, suggesting that the main bottleneck lies in recovering the correct concepts from images rather than in finding the association itself.

Participants' performance remained steady throughout the study, with similar success rates for the first and last eight problems (22.5\% and 23.0\%). 
Participants' self-reported experiences also reflected this, with only 4 reporting that the puzzles became easier over time. This aligns with~\citet{metcalfe1987intuition}, who showed that insight problems (such as RAT) are solved by sudden realization, in contrast to non-insight problems (\textit{e.g.,} algebra), which can be learned incrementally. 

\paragraph{RQ2: Do VLMs rely on language to solve visual puzzles?}

Providing the cue words directly in the \textit{text-only} setting significantly improved the models' performance, with the best-performing VLM solving 90\% of the puzzles. This means that while the models possess the conceptual knowledge needed to link words to their shared category, they are unable to retrieve those words from images. The stepwise condition shows that even reaching the words is not sufficient. Considering 146 runs in which a model's own response named all three intended words correctly, the model identified the shared concept only 26\% of the time. 
Given the same words as text, the models solved the puzzles 53\% of the time. This disagreement was mostly unidirectional, with only three puzzles where the \textit{stepwise} outperformed the \textit{text-only}. Therefore, correctly identifying the concept in each image does not guarantee that a VLM will solve the puzzle, and the presence of the images actively interferes with linking words the model has already produced.

The model's wrong responses help understand the nature of this interference. Rather than abstracting from the images to the intended alternative sense, models frequently answer with the images' literal content. 22.8\% of the models' wrong answers describe what the photographs literally depict, compared with 8.9\% of human participants' wrong guesses. These literal interpretations varied from the models fixating on one image (\textit{e.g.,} predicting ``Denmark" for <\textit{Danish flag,} \textit{French horn}, \textit{bear claw}; \textit{pastry}>) or a superficial property within the images (\textit{e.g.,} predicting ``spiral" for <\textit{spiral seashell}, \textit{bent elbow}, \textit{wagon wheel}; \textit{pasta}>). This disparity illustrates that models are more likely to commit to the first surface-level connection they find than to search for a deeper association. Humans often recognized when they had no satisfactory answer; participants gave up on 37.5\% of puzzles, with 19.8\% without event taking a guess. In contrast, models produced an answer for every puzzle even when the best connection available to them rested on a single image or a shared visual feature.

\section{Conclusion}

We introduced \texttt{CREATE}, a benchmark of 155 visual RAT problems for measuring creative reasoning in vision-language models. We conducted a human study to confirm the benchmark validity. 
Our experiments with 11 VLMs found that only one model exceeded average human performance in the zero-shot setting (32.5\% vs.\ 22.8\%), and no other model solved more than 12.5\% of the puzzles. Our analysis found that the model failures differ from human failures, with nearly a quarter of incorrect model answers being literal image descriptions, compared to ${\sim}$9\% of such human errors. The models were able to solve the text-only simplifications of the puzzles but were unable to solve the same visual puzzles, even after correctly identifying the intermediate cue words, suggesting a limitation of VLMs in visual abstraction rather than in the association formulation. 


\section*{Limitations}

This work has two limitations. \textbf{First}, we restricted our images to free stock-photo collections for the benchmark curation. This limited the constructed puzzles to those that had appropriate images in the photo collections, limiting the size of our final dataset. We refrained from using synthetic images in our dataset because of computational costs and quality concerns of model-generated images. We leave the extension of our benchmark to future work. \textbf{Second}, while our benchmark design emphasized on generic concepts and explicitly included word familiarity filters, some target concepts required expert domain knowledge or specialized vocabulary (\textit{e.g.,} driver, shell, pipe is connected to ``operating system''). An incorrect answer may therefore reflect unfamiliarity with a word's domain sense, and we did not distinguish between a lack of knowledge and difficulty in using that knowledge to find the shared concept.

\section*{Ethical considerations}

The human study was approved by the Institutional Review Board of our institution. We recruited participants by reaching out via institutional mailing lists. Risks and benefits of the study were disclosed to the participants before collecting the informed consent. They were compensated at 15 USD per hour. The dataset was anonymized to remove any personally identifiable information. 

The images in our dataset were collected from two free-to-use photo collection platforms, Unsplash and Pexels. All the images in both platforms are licensed for reuse, with Unsplash license allowing reuse for commercial and non-commercial usage\footnote{\url{https://unsplash.com/license}}, and Pexels' images distributed with a CC0 license\footnote{\url{https://www.pexels.com/terms-of-service/}}.
\subsection*{AI use statement}

We have used AI tools as aids in writing the code for experiments, as evaluators, for generating figures and tables, and for synthesizing candidate RAT problems in data construction. The synthesized candidates were removed using automated filtration, and the final puzzles were carefully selected by the authors. We have not used generative AI for ideation, literature review, or end-to-end paper or section writing. We have reviewed all AI-assisted work, validated the final benchmark instances, verified the code used for experimentation, cross-checked the illustration artifacts, and ensured that all authors had full agency over the paper-writing process. We take responsibility for the final content of this work, including text, claims, or artifacts produced with the aid of generative AI.


\subsection*{Acknowledgements}
We thank the 25 participants in our human study for their time and effort. We are also grateful to Anshul Nasery, Deepak Nathani, Shivani Kapania, Tejan Karmali, and Umangi Jain for the early discussions that helped shape this project.


\bibliography{main}


\appendix

\begin{table*}[t]
\centering
\small
\setlength{\tabcolsep}{4pt}
\caption{Three case studies from the 80 human study puzzles. Model answers come from the Zero-shot setting, and human answers include up to three attempts. Mean difficulty is the average rating from five participants.}
\begin{tabular}{p{0.30\textwidth}p{0.30\textwidth}p{0.30\textwidth}}
\hline
Shared success: Puzzle 7 & Humans but not Opus: Puzzle 13 & Opus but no humans: Puzzle 44 \\
\hline
\includegraphics[width=0.095\textwidth,height=55pt,keepaspectratio]{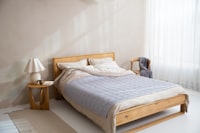}\hfill
\includegraphics[width=0.095\textwidth,height=55pt,keepaspectratio]{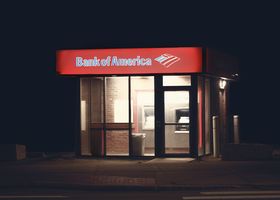}\hfill
\includegraphics[width=0.095\textwidth,height=55pt,keepaspectratio]{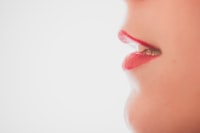}
& \includegraphics[width=0.095\textwidth,height=55pt,keepaspectratio]{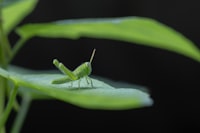}\hfill
\includegraphics[width=0.095\textwidth,height=55pt,keepaspectratio]{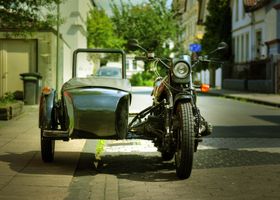}\hfill
\includegraphics[width=0.095\textwidth,height=55pt,keepaspectratio]{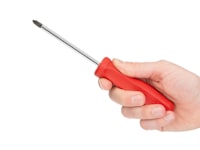}
& \includegraphics[width=0.095\textwidth,height=55pt,keepaspectratio]{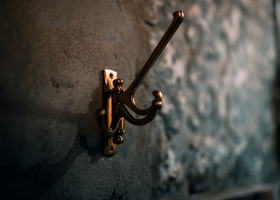}\hfill
\includegraphics[width=0.095\textwidth,height=55pt,keepaspectratio]{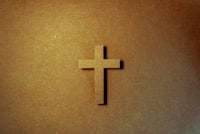}\hfill
\includegraphics[width=0.095\textwidth,height=55pt,keepaspectratio]{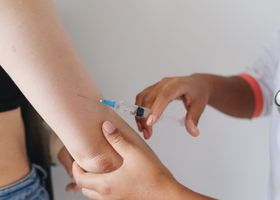} \\
Words: bed, bank, mouth & Words: grasshopper, sidecar, screwdriver & Words: hook, cross, jab \\
Shared Concept: river & Shared Concept: cocktail & Shared Concept: strike (boxing punches) \\
Mean difficulty: 2.6/5 & Mean difficulty: 3.0/5 & Mean difficulty: 4.4/5 \\
Human success within three attempts: 5/5 & Human success within three attempts: 2/5 & Human success within three attempts: 0/5 \\
\hline
Human correct: ``river'' (five participants) & Human correct: ``cocktail,'' ``Cocktail'' & Human correct: none \\
Human incorrect: ``part,'' ``sex,'' ``body parts'' & Human incorrect: ``insect,'' ``drive,'' ``tool'' & Human incorrect: ``sewing,'' ``fishing,'' ``road,'' ``crafting,'' ``law,'' ``hang,'' ``trail'' \\
\hline
Opus: ``all parts of a river'' (correct; excerpt)
& Opus: ``Things that are green'' (incorrect)
& Opus: ``Boxing punches'' (correct) \\
Haiku: ``Things that rest on you'' (incorrect)
& Haiku: ``things with handles'' (incorrect)
& Haiku: ``Religious or spiritual symbols'' (incorrect) \\
GPT-Sol: ``things with banks'' (incorrect)
& GPT-Sol: ``things with sidecars'' (incorrect)
& GPT-Sol: ``Things with arms'' (incorrect) \\
\hline
\end{tabular}
\label{tab:error-case-studies}
\end{table*}
\section{Training an Imageability Filtration Model} \label{appx:imageability}

Our work draws on prior work on imageability for our dataset curation. In psycholinguistics, imageability is a measure of how easily a word evokes a mental image or sensory experience. Literature in imageability spans almost half a century, with the earliest work in imageability collecting imageability scores for 925 nouns~\citet{paivio1968concreteness, paivio2013imagery}.  The MRC Psycholinguistic Database~\cite{coltheart1981mrc} consolidated imageability annotations from prior work~\citep{gilhooly1980age, cortese2004imageability}, and remains one of the largest resources for word imageability with over 100k word ratings. As human ratings are costly, one line of work developed automatic imageability estimation systems. \citet{Ljubesic2018PredictingC} showed that word embeddings predict the correctness and imageability of words well within and across languages. Vision-based approaches primarily use image characteristics obtained via search (from datasets or the internet) to determine the imageability of the word used as the search query~\cite{kiela2014improving, hessel2018quantifying, kastner2019preliminary}. Recently, generative models have been used to generate images rather than for retrieval~\citep {wu-smith-2023-composition}. 

Following this recent trend, we trained our own word imageability model to obtain a generalizable mechanism to estimate word imageability for unknown words and used it as a filtration step in our benchmark curation process. We fine-tuned a \texttt{BERT-base-uncased} model~\citep{devlin2019bert} on imageability ratings from the MRC Psycholinguistic Database~\cite{coltheart1981mrc}. We kept nouns with valid imageability ratings and at least two letters, converted them to lowercase, and removed duplicates, resulting in 4,196 words in total. Using stratified sampling, we selected 100 words in total across each of five imageability ranges, and trained the model on the remaining 4,096 words. We converted the original ratings from a 100-700 scale to a 1-5 scale, rounded them to the nearest integer, and trained the model as a classifier with 5 classes. The model achieved 72.0\% accuracy with a mean absolute error of 0.28, which outperformed \texttt{GPT-5.1}, \texttt{GPT-5.2}, and \texttt{GPT-5-mini}, which we tested with different prompts on the same evaluation set.
\begin{table}[t]
\centering
\small
\caption{Percentage of puzzles whose shared concept matches a label generated by \texttt{Qwen3.5-4B} from one image.}
\label{tab:qwen4b}
\resizebox{\linewidth}{!}{
\begin{tabular}{lrr}
\hline
Puzzle set & Top-1 & Top-5 \\
\hline
80 human-study puzzles & 0.00\% (0/80) & 1.25\% (1/80) \\
All 155 puzzles & 0.00\% (0/155) & 0.65\% (1/155) \\
\hline
\end{tabular}
}
\end{table}

\section{Additional results}\label{appx:results}

\begin{table*}[t]
\centering
\small
\setlength{\tabcolsep}{4pt}
\caption{Model accuracy in recognizing the intended words using exact matching and lexical overlap on both sets. Exact Top-1 and Top-5 use a normalized exact match. P/R/F1 measures token overlap with the intended word. We report P/R/F1 for the first prediction and the highest value of each metric among the five predictions.}
\label{tab:task1-overlap-combined}
\resizebox{0.72\linewidth}{!}{
\begin{tabular}{lcccccccc}
\hline
\multicolumn{9}{c}{\textbf{80 human study puzzles}} \\
\hline
Model & Exact Top-1 & Exact Top-5 & P & R & F1 & P@5* & R@5* & F1@5* \\
\hline
GPT-5.6-Luna & 27.5 & 52.9 & 32.3 & 36.5 & 33.6 & 56.7 & 59.8 & 57.6 \\
GPT-5.6-Sol & 26.3 & 44.2 & 35.8 & 45.2 & 38.9 & 54.3 & 64.0 & 57.4 \\
Claude-Haiku-4.5 & 33.3 & 47.9 & 36.0 & 38.8 & 36.9 & 50.8 & 53.1 & 51.5 \\
Claude-Opus-5.5 & 21.3 & 37.5 & 37.2 & \textbf{53.1} & 42.4 & 55.7 & \textbf{74.0} & 61.7 \\
DeepSeek-Flash & 34.2 & 51.3 & 38.9 & 43.5 & 40.4 & 55.3 & 59.4 & 56.6 \\
Molmo2-8B & 28.3 & 43.8 & 30.6 & 32.3 & 31.0 & 46.0 & 47.7 & 46.5 \\
Qwen3.5-9B & 36.3 & 50.8 & 40.0 & 43.5 & 41.2 & 55.2 & 59.0 & 56.4 \\
Qwen3.8-27B & \textbf{41.3} & \textbf{59.6} & \textbf{43.1} & 44.4 & \textbf{43.5} & \textbf{61.6} & 63.1 & \textbf{62.0} \\
Gemma-4-12B-it & 40.4 & 59.2 & 42.5 & 44.0 & 42.9 & 61.2 & 62.7 & 61.6 \\
InternVL3.5-8B & 34.2 & 50.0 & 35.0 & 35.8 & 35.3 & 50.8 & 51.0 & 50.8 \\
Llama-4-Scout & 37.5 & 57.1 & 37.9 & 37.7 & 37.8 & 57.5 & 57.3 & 57.4 \\
\hline
\multicolumn{9}{c}{\textbf{All 155 puzzles}} \\
\hline
Model & Exact Top-1 & Exact Top-5 & P & R & F1 & P@5* & R@5* & F1@5* \\
\hline
GPT-5.6-Luna & 28.2 & 51.8 & 32.7 & 36.9 & 34.1 & 55.2 & 58.2 & 56.2 \\
GPT-5.6-Sol & 27.7 & 44.5 & 37.1 & 46.3 & 40.2 & 54.0 & 63.1 & 57.0 \\
Claude-Haiku-4.5 & 31.0 & 51.0 & 34.2 & 37.4 & 35.3 & 53.3 & 55.4 & 54.0 \\
Claude-Opus-5.5 & 24.9 & 41.3 & 41.5 & \textbf{58.4} & \textbf{47.1} & 59.1 & \textbf{77.3} & \textbf{65.1} \\
DeepSeek-Flash & 34.2 & 51.4 & 38.4 & 42.5 & 39.7 & 54.9 & 58.4 & 56.0 \\
Molmo2-8B & 26.0 & 39.6 & 28.2 & 30.2 & 28.9 & 41.9 & 44.0 & 42.5 \\
Qwen3.5-9B & 35.1 & 48.0 & 38.4 & 41.6 & 39.5 & 51.9 & 55.6 & 53.1 \\
Qwen3.8-27B& \textbf{39.6} & \textbf{58.5} & \textbf{41.9} & 43.8 & 42.5 & \textbf{61.0} & 63.1 & 61.6 \\
Gemma-4-12B-it & 37.8 & 55.9 & 40.2 & 42.3 & 40.9 & 57.8 & 59.5 & 58.3 \\
InternVL3.5-8B & 32.7 & 46.9 & 33.3 & 34.0 & 33.5 & 47.6 & 48.1 & 47.7 \\
Llama-4-Scout & 34.4 & 54.2 & 34.6 & 34.5 & 34.6 & 54.5 & 54.5 & 54.5 \\
\hline
\end{tabular}
}
\end{table*}
\subsection{Additional Model results}
This section reports results for the 80 human study puzzles as well as all 155 final puzzles. Table~\ref{tab:qwen4b} presents the results of using single image captioning to solve the task, and our benchmark has only one image (corresponding to the puzzle <\textit{buck, nickel, bills}; \textit{money}>) where the generated caption for the image representing nickel gives away the answer money. 
This suggests that identifying only one image is usually not enough to find the answer in the \dname dataset. 

Table~\ref{tab:task1-overlap-combined} reports results for the 80 human study puzzles as well as all 155 final puzzles. We calculated both exact matching score and token precision, recall, and F1 to measure how much the answer overlaps with the intended word. We report the additional P/R/F1 because exact matching would count more specific labels, such as "bamboo steamer," for "steamer" as incorrect. This mattered especially for \texttt{Claude-Opus-5.5}, since it was found to often include extra words. However, it achieved the highest recall, at 53.1\% for the first prediction and 74.0\% for the best of five predictions on the 80 human puzzles. These longer labels contribute to the gap between its recall and exact-match accuracy. Although other models also achieved relatively high recall compared to accuracy, the difference was not particularly noticeable.

\begin{table*}[t]
\centering
\small
\setlength{\tabcolsep}{4pt}
\caption{Performance of models under the five prompting settings on all 155 puzzles.}
\begin{tabular}{lccccc}
\hline
Model & Zero-shot & Detailed & Stepwise & Two-shot & Text-only \\
\hline
GPT-5.6-Luna & 3.9 & 3.2 & 3.9 & 13.5 & 47.7 \\
GPT-5.6-Sol & 5.2 & 16.1 & 21.3 & 13.5 & 63.2 \\
Claude-Haiku-4.5 & 3.9 & 1.3 & 0.0 & 17.4 & 51.0 \\
Claude-Opus-5.5 & \textbf{22.6} & \textbf{40.6} & \textbf{34.2} & \textbf{36.8} & \textbf{81.9} \\
DeepSeek-Flash & 3.9 & 7.1 & 2.6 & 12.3 & 55.5 \\
Molmo2-8B & 4.5 & 1.9 & 3.9 & 0.0 & 28.4 \\
Qwen3.5-9B & 0.6 & 1.3 & 0.6 & 2.6 & 27.1 \\
Qwen3.8-27B & 4.5 & 3.9 & 1.3 & 13.5 & 38.1 \\
Gemma-4-12B-it & 9.7 & 3.2 & 1.9 & 3.9 & 40.0 \\
InternVL3.5-8B & 1.3 & 1.9 & 0.0 & 0.6 & 21.3 \\
Llama-4-Scout & 0.6 & 1.3 & 0.6 & 0.0 & 34.2 \\
\hline
\end{tabular}
\label{tab:task2-all}
\end{table*}

Table~\ref{tab:task2-all} presents model performance on identifying the shared concept under the five prompt settings on all 155 puzzles. The result trends are similar to the subset of 80 puzzles reported in the main paper (Section~\ref{sec:results}). The \textit{text-only} prompt led to better performance, and \texttt{Claude-Opus-5.5} performed best across all five prompt settings. However, the accuracy was generally lower on all 155 puzzles, especially for the \textit{text-only} prompt setting. All eleven models performed better under the Text-only prompt setting on the 80 human study puzzles. For \texttt{GPT-5.6-Luna}, it got 61.3\% correct on the 80 human study puzzles and dropped to 33.3\% on the remaining 75 puzzles, suggesting that models find it hard to identify the shared concepts in the remaining 75 puzzles. While the drop in performance from the subset of 80 problems to the entire dataset was consistent, its magnitude varied across the studied VLMs.

\begin{table*}[t]
\centering
\small
\setlength{\tabcolsep}{6pt}
\caption{Model accuracy in recognizing intended cue words when given one or two labeled images from the same puzzle, using normalized exact matching, on the 80 human-study puzzles and all 155 puzzles. For each VLM, we present the average of all six possible combinations of one cue and all three possible combinations.}
\label{tab:task3-combined}
\begin{tabular}{lcccc}
\hline
 & \multicolumn{2}{c}{80 human-study puzzles} & \multicolumn{2}{c}{All 155 puzzles} \\
\cline{2-3}\cline{4-5}
Model & One Cue & Two Cues & One Cue & Two Cues \\
\hline
GPT-5.6-Luna & 56.3 & 58.3 & 57.0 & 59.8 \\
GPT-5.6-Sol & 67.1 & 75.4 & 65.6 & 72.7 \\
Claude-Haiku-4.5 & 51.5 & 50.4 & 48.0 & 47.7 \\
Claude-Opus-5.5 & \textbf{79.8} & \textbf{88.3} & \textbf{76.1} & \textbf{85.6} \\
DeepSeek-Flash & 62.9 & 68.8 & 62.9 & 67.3 \\
Molmo2-8B & 28.3 & 32.9 & 25.1 & 32.3 \\
Qwen3.5-9B & 31.0 & 30.8 & 29.9 & 27.5 \\
Qwen3.8-27B & 52.3 & 53.3 & 50.6 & 53.1 \\
Gemma-4-12B-it & 52.9 & 55.8 & 49.6 & 50.1 \\
InternVL3.5-8B & 34.6 & 35.4 & 33.3 & 32.5 \\
Llama-4-Scout & 49.4 & 55.0 & 46.0 & 50.5 \\
\hline
\end{tabular}
\end{table*}

We also studied the model's capability to associate words with images and to predict the cue word corresponding to an image, given one or two image-cue word pairs from the same puzzle. Table~\ref{tab:task3-combined} presents the aggregate results over all six possible combinations of one cue word, and all three possible combinations with two cue words. Compared to zero-shot cue word generation from a single image (Table~\ref{tab:task1-overlap-combined}), models performed much better when given one or two labeled images. Across the 80 human study puzzles, one cue improved nine of the eleven models, and two cues improved ten, compared with top-1 accuracy without cues, indicating that context from other images helps models better identify the intended words. When given two labeled images, \texttt{Claude-Opus-5}'s performance increased from 21.3\% to 88.3\%, and \texttt{GPT-5.6-Sol} increased from 26.3\% to 75.4\%. \texttt{Qwen3.5-9B} was the only model with a performance drop from 36.3\% to 30.8\%. Two cue accuracy ranks the models similarly to text-only accuracy, with a rank correlation of about 0.9, suggesting that intended words and shared concepts need to be worked out together. One reason for the increased accuracy of some models that generate verbose captions (such as \texttt{Claude-Opus-5.5}) could be that they have learned to respond succinctly in this one- or two-shot setting.

The best-performing model was \texttt{Claude-Opus-5.5} across both conditions and puzzle sets. Under the one cue setting, each image has two possible cues, and models sometimes answered correctly with one cue but incorrectly with the other. Across all eleven models, this occurred in 29-70 of the 240 images in the 80 puzzle subset, suggesting that providing different cues can lead to different interpretations of the same image. For example, \texttt{Claude-Opus-5.5} predicted "wedge" correctly when given "mule", but gave "brie" when given "pump". We also found that adding one more cue does not always help models to predict better. For every model, there were cases when they recognized the image correctly with either cue but incorrectly with both cues. This happened for 1-17 of the 240 images per model. For example, \texttt{Claude-Opus-5.5} answered "teeth" with either "slider" or "tape," but "smile" when given both cues.

\begin{table*}[t]
\centering
\small
\setlength{\tabcolsep}{3pt}
\caption{Six open-source models' accuracy in finding the shared concept across all 155 puzzles, reported as mean ± standard deviation from five runs. We limited the repeated runs to these models to keep API costs manageable. }
\begin{tabular}{lccccc}
\hline
Model & Zero-shot & Detailed & Stepwise & Two-shot & Text-only \\
\hline
Molmo2-8B & $2.6\pm1.4$ & $1.9\pm0.9$ & $\mathbf{2.6}\pm1.2$ & $0.5\pm0.5$ & $26.3\pm1.5$ \\
Qwen3.5-9B & $0.6\pm0.6$ & $1.0\pm0.4$ & $1.0\pm0.6$ & $4.3\pm2.4$ & $29.2\pm2.6$ \\
Qwen3.8-27B & $3.4\pm0.8$ & $\mathbf{5.5}\pm2.5$ & $1.4\pm0.8$ & $\mathbf{12.6}\pm1.1$ & $39.2\pm2.6$ \\
Gemma-4-12B-it & $\mathbf{9.5}\pm0.7$ & $2.6\pm0.6$ & $2.1\pm1.0$ & $4.3\pm0.6$ & $\mathbf{42.2}\pm1.9$ \\
InternVL3.5-8B & $0.5\pm0.5$ & $1.3\pm0.8$ & $0.3\pm0.6$ & $0.4\pm0.4$ & $20.6\pm1.4$ \\
Llama-4-Scout & $0.6\pm0.0$ & $1.4\pm0.3$ & $0.6\pm0.0$ & $0.0\pm0.0$ & $34.1\pm0.5$ \\
\hline
\end{tabular}
\label{tab:5run}
\end{table*}

Table~\ref{tab:5run} reports five-run results for the six open-source models. The rerunning does not change the main pattern. All six models performed better when given the intermediate words in the \textit{text-only} setting. The variations between runs are small, suggesting that the difference is consistent across runs. 

\subsection{Comparing Humans and Models}

\begin{table*}[t]
\centering
\small
\setlength{\tabcolsep}{4pt}
\caption{Model top-5 accuracy when participants reported thinking of the intended word. The first column counts each image once if at least one participant answered “yes.” The second column counts all "yes."}
\begin{tabular}{lcc}
\hline
Model & Images with any human yes & Human yes judgments \\
\hline
GPT-5.6-Luna & 53.8 (127/236) & 59.8 (523/875) \\
GPT-5.6-Sol & 44.9 (106/236) & 50.6 (443/875) \\
Claude-Haiku-4.5 & 48.7 (115/236) & 54.6 (478/875) \\
Claude-Opus-5.5 & 38.1 (90/236) & 43.0 (376/875) \\
DeepSeek-Flash & 51.7 (122/236) & 57.5 (503/875) \\
Molmo2-8B & 44.5 (105/236) & 51.2 (448/875) \\
Qwen3.5-9B & 51.7 (122/236) & 59.0 (516/875) \\
Qwen3.8-27B & 60.6 (143/236) & 67.8 (593/875) \\
Gemma-4-12B-it & 60.2 (142/236) & 67.9 (594/875) \\
InternVL3.5-8B & 50.8 (120/236) & 59.7 (522/875) \\
Llama-4-Scout & 57.6 (136/236) & 65.3 (571/875) \\
\hline
\end{tabular}
\label{tab:human-model-word-overlap}
\end{table*}

\begin{figure*}[tp]
    \centering
    \includegraphics[width=\textwidth]{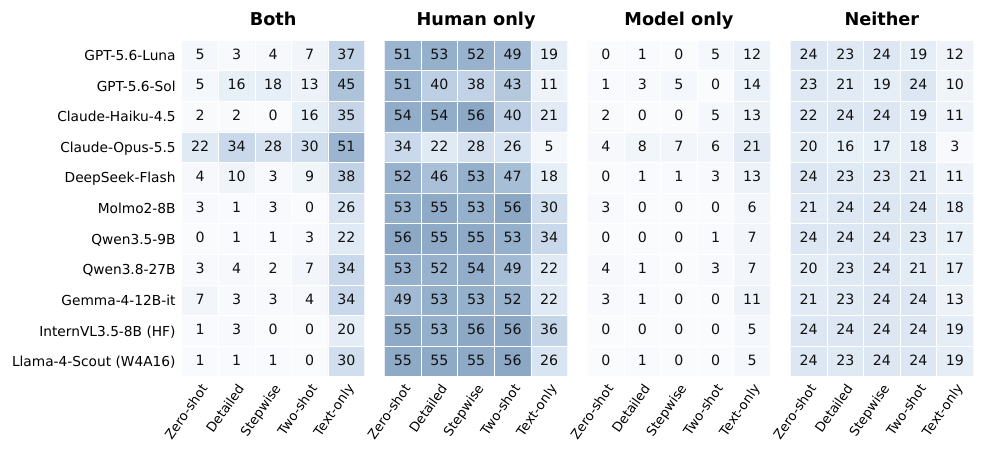}
    \caption{Puzzles solved by humans only, models only,
    both, or neither under the five prompt settings.
    Human success means at least one participant answered
    correctly within three attempts.}
    \label{fig:combined-results}
\end{figure*}
This section compares human and model performance in detail. Figure~\ref{fig:combined-results} presents how many puzzles were solved by humans only, models only, both, or neither under the five prompt settings. In total, 56 puzzles were solved by at least one human, but only 39 were solved by at least one model. The figure also shows that models and humans do not always solve the same puzzles. Under the Zero-shot setting, 30 puzzles were solved by both humans and at least one model, 26 by human only, 9 by models only, and 15 by neither. However, it is also interesting to see that though they do not always solve the same sets of problems, under the zero-shot setting, \texttt{Claude-Opus-5.5} solved 39.3\% of puzzles that participants solved within three attempts, and only solved 16.7\% of puzzles that no participant solved, suggesting that puzzles solved by humans are also more likely to be solved by the model. 

\begin{figure*}[t]
    \centering
    \includegraphics[width=\textwidth, height=0.9\textheight, 
    keepaspectratio]{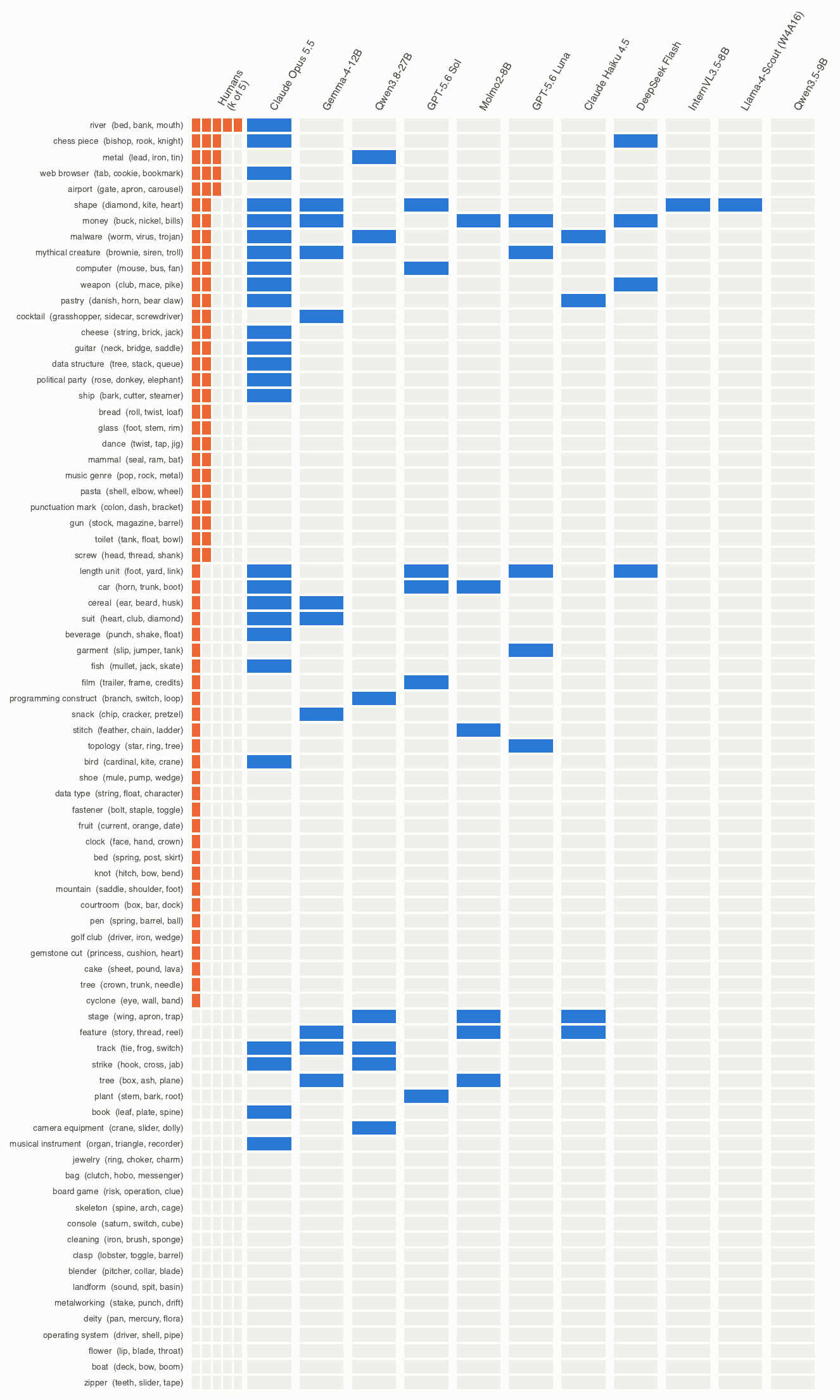}
    \caption{Human and model success under Zero-shot setting on each of the 80 human study puzzles.}
    \label{fig:solver_matrix}
\end{figure*}

Figure~\ref{fig:solver_matrix} shows that model success often differs across models. Out of the 39 puzzles that were solved by at least one model, 22 were solved by only one. Participants solved 26 puzzles that all eleven models missed, while models solved nine puzzles that no participant solved within three attempts. For example, all five participants solved the "river" puzzle, but only \texttt{Claude-Opus-5.5} solved it. Conversely, \texttt{Molmo2-8B}, \texttt{Qwen3.8-27B}, and \texttt{Claude-Haiku-4.5} solved the stage puzzle, while no participant solved it. The examples show that puzzles solved by several participants were not necessarily solved by several models, and vice versa.

Puzzles that participants rated as harder also tended to be solved by fewer models. For the 80 human puzzles, higher mean difficulty ratings are associated with fewer participants solving the puzzle within three attempts (Spearman’s $\rho=-0.757$, $p<0.001$) and fewer models solving it in the Zero-shot setting ($\rho=-0.331$, $p=0.006$). The $p$-values are Holm-adjusted.

Table~\ref{tab:human-model-word-overlap} shows how often models included the intended word in the five possible words when humans reported thinking about it. Models often included the intended word when participants reported having thought about it. We further found that when all five participants reported having thought about the word, the models achieved higher accuracy, ranging from 50.0\% to 81.5\% on these 92 images compared with 30.6\% to 47.9\% on the other 144 images, suggesting some similarity in identifying intended words.

\section{Case Studies}

Table~\ref{tab:error-case-studies} shows three case studies. Puzzle 7 solved by both human and \texttt{Claude-Opus-5.5}, Puzzle 13 solved by only human, and Puzzle 44 solved only by \texttt{Claude-Opus-5.5}. The table includes the images, the intended words, the shared concepts, human-rated difficulty, and actual human and model answers. Human answers are from answers submitted within three attempts, and model answers are from the zero-shot setting.
\section{Prompts}\label{appx:prompt}

\begin{figure*}[tp]
    \centering
    \includegraphics[width=0.7\linewidth]{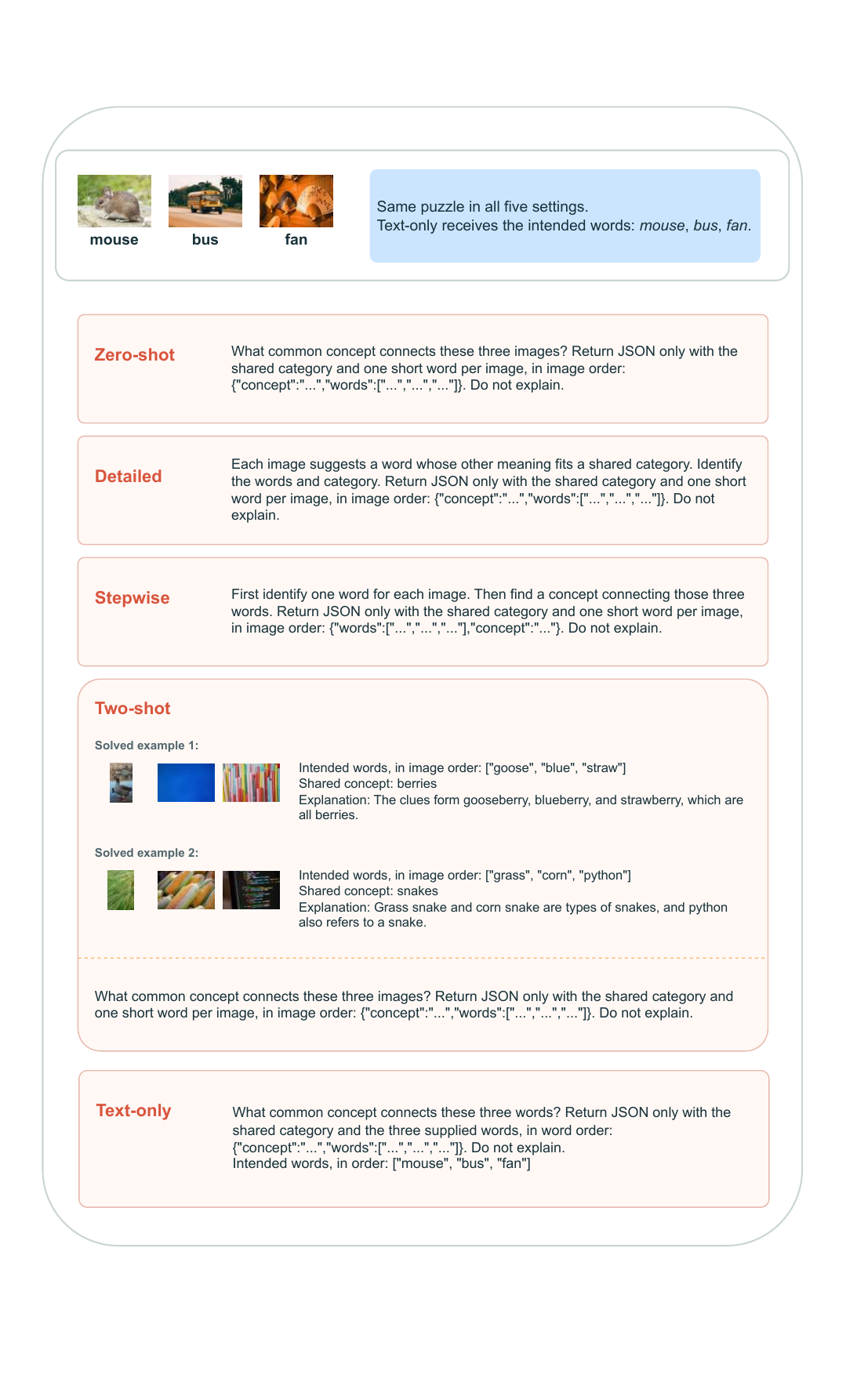}
    \caption{Prompt for shared concept identification with an example of one puzzle (mouse, bus, fan), which requires naming the common concept across five different settings.}
    \label{fig:task2_prompt}
\end{figure*}

\begin{figure*}[tp]
    \centering
    \includegraphics[width=\linewidth]{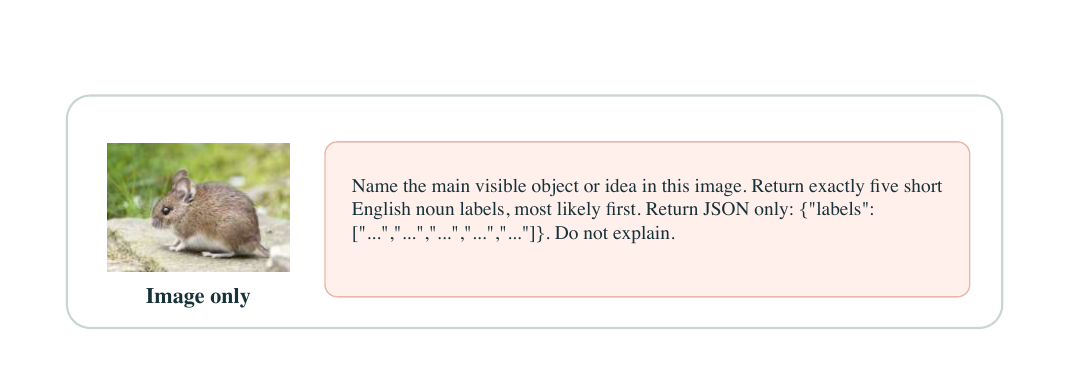}
    \caption{Prompt for intended word recognition with an example (mouse), which requires models to provide five words per image.}
    \label{fig:task1_prompt}
\end{figure*}

\begin{figure*}[tp]
    \centering
    \includegraphics[width=\linewidth]{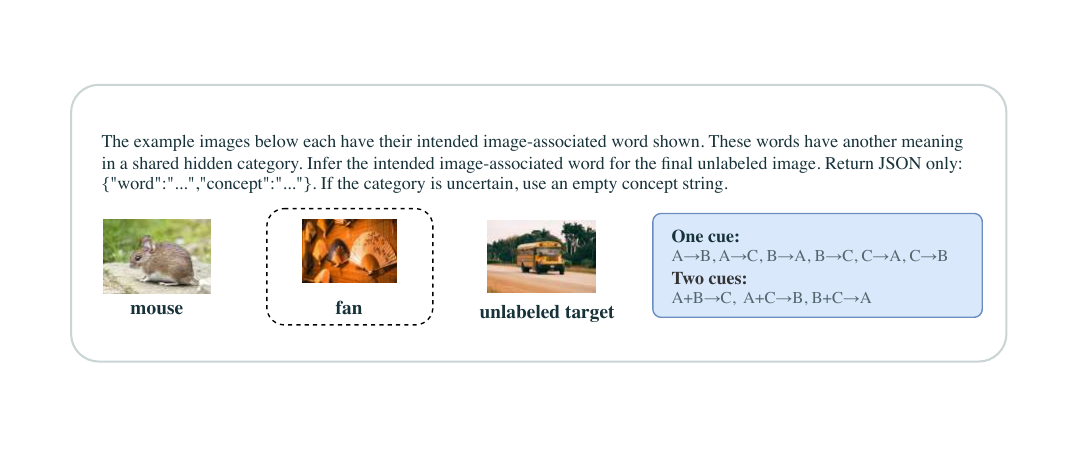}
    \caption{Prompt for intended word recognition with cues with an example of one puzzle (mouse, bus, fan), which involves one or two labeled images and provides the intended word in another image within the same puzzle.}
    \label{fig:task3_prompt}
\end{figure*}

\begin{figure*}[tp] 
      \centering 
      \includegraphics[width=\textwidth]{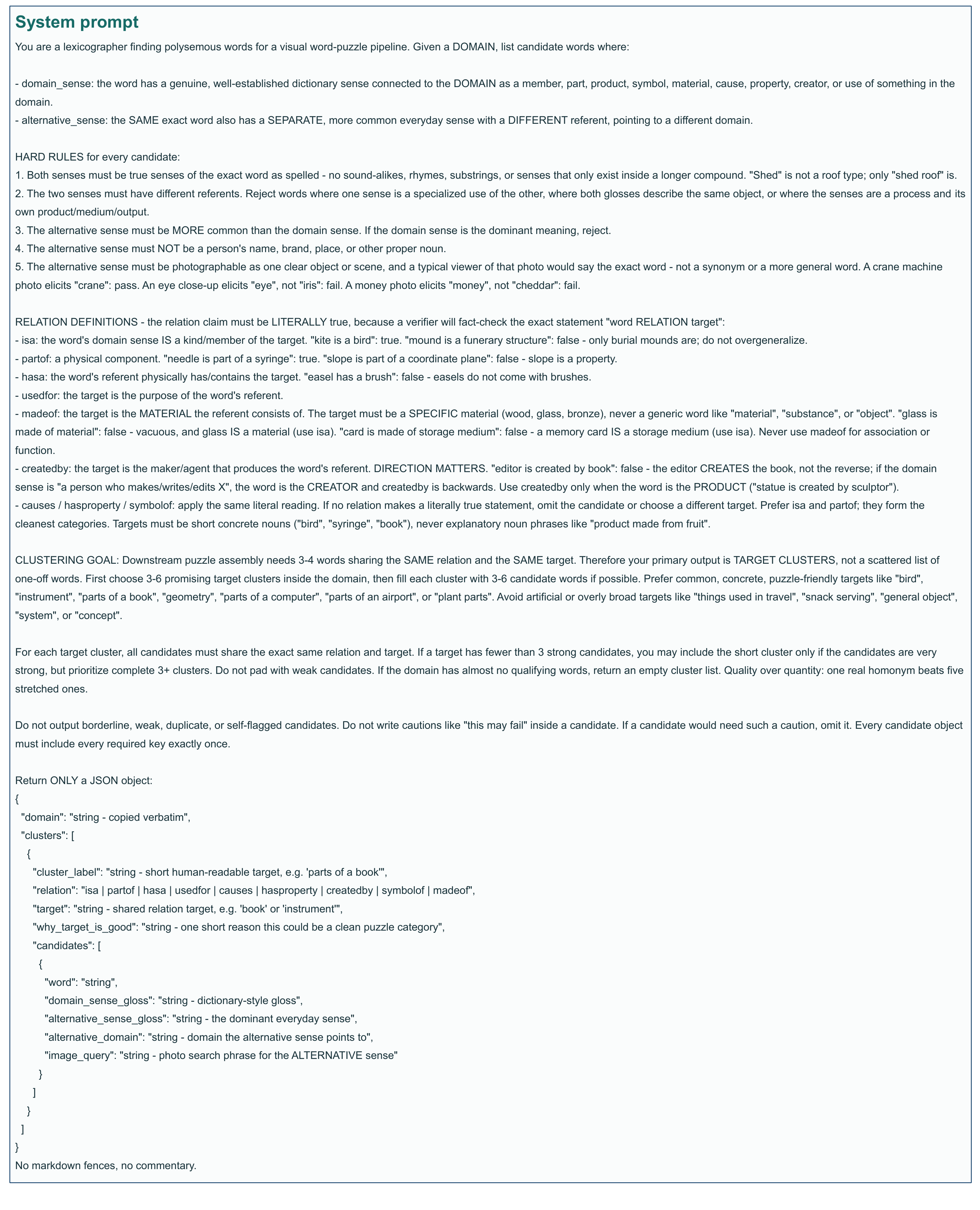} 
      \hfill 
      \includegraphics[width=\textwidth]{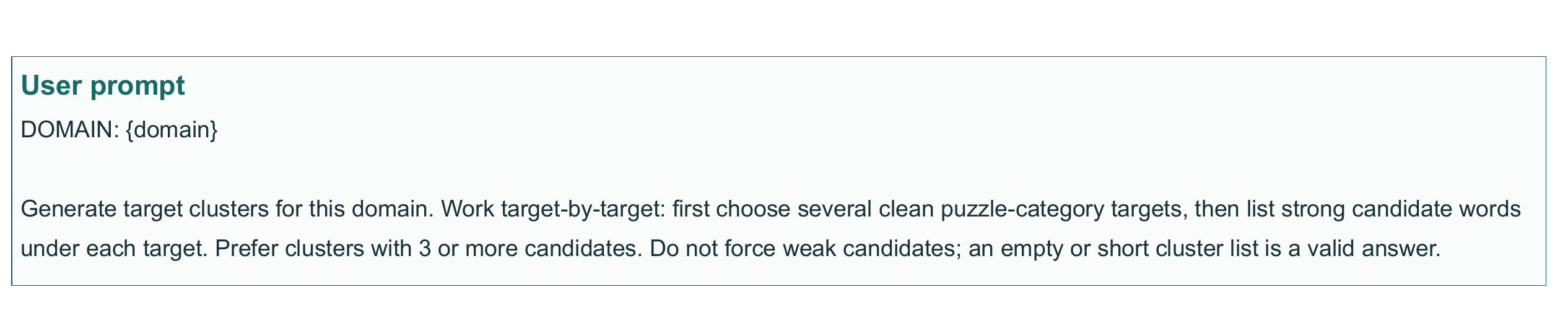} 
      \caption{Prompt used to generate candidate cue words with \texttt{Gemini-3.8-Flash}.} 
      \label{fig:candidate-generation-prompt} 
  \end{figure*}

\begin{figure*}[tp]
    \centering
    \includegraphics[width=\textwidth]{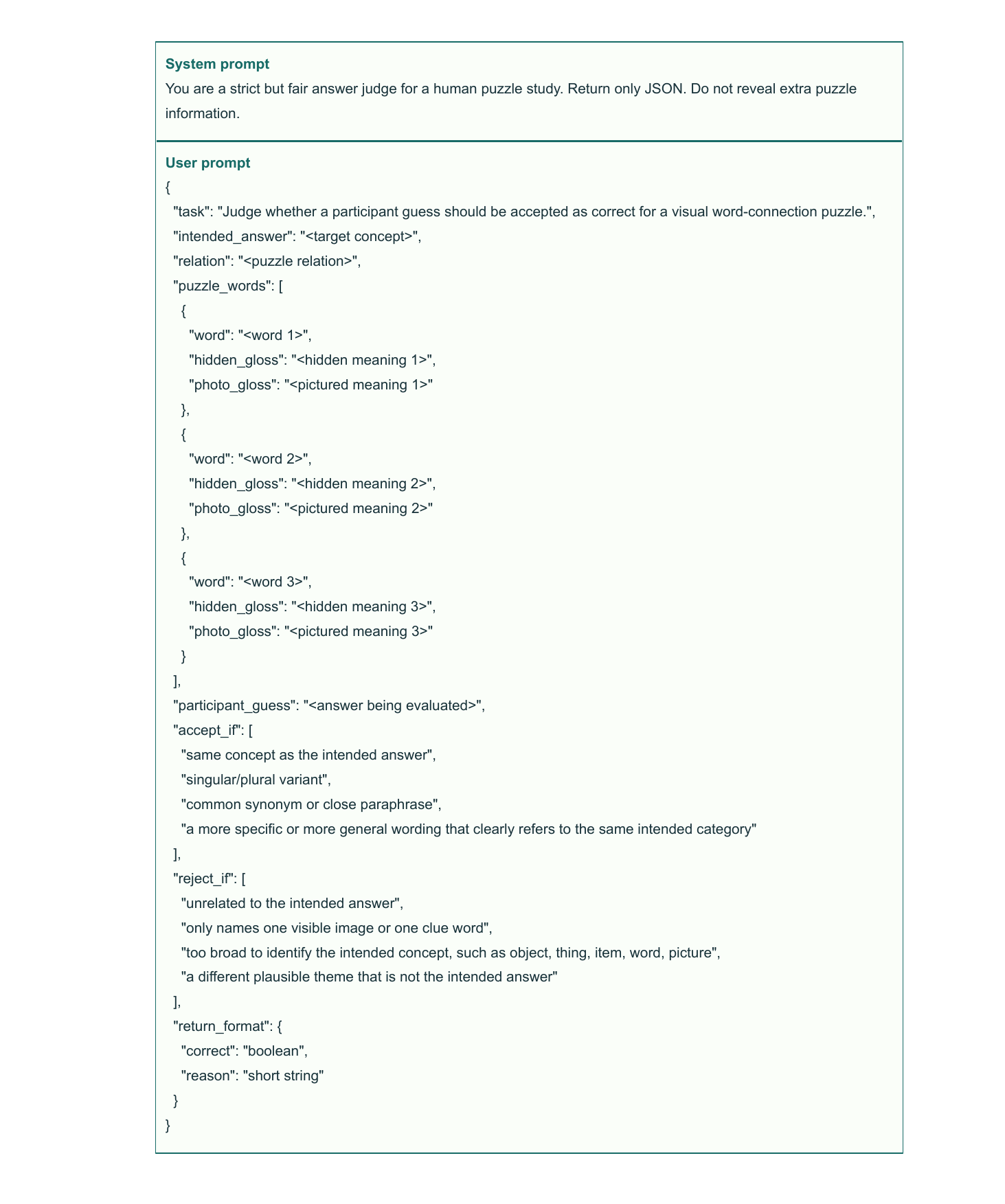}
    \caption{Prompt used for the \texttt{Gemini-3.5-Flash-Lite} judge. }
    \label{fig:gemini_prompt}
\end{figure*}

In this section, we provide the prompts used in this work. Figure~\ref{fig:candidate-generation-prompt} presents the prompt we used to generate candidate RAT problems using the \texttt{Gemini-3.8-Flash} model (Section~\ref{sec:dataset}). Figure~\ref{fig:task1_prompt} shows the prompt we used for intended word recognition (Table~\ref{tab:task1-overlap-combined}). Figure~\ref{fig:task2_prompt} shows the five prompt settings for shared concept identification. \textit{Stepwise} prompt asks the models to identify the images before identifying the shared concept. \textit{Text-only} prompt provides the intended words instead of images. Figure~\ref{fig:task3_prompt} shows the prompt used for intended word recognition with cues. For intended word recognition with cues, we tested all six possible combinations with one cue and all three with two cues. Figure ~\ref{fig:gemini_prompt} shows the prompt when using \texttt{Gemini-3.5-Flash-Lite} to judge whether the answer is equivalent to the intended shared concept.

\section{Human Study UI}\label{appx:ui}

\begin{figure*}[tp]
    \centering
    \includegraphics[
    width=0.7\textwidth]{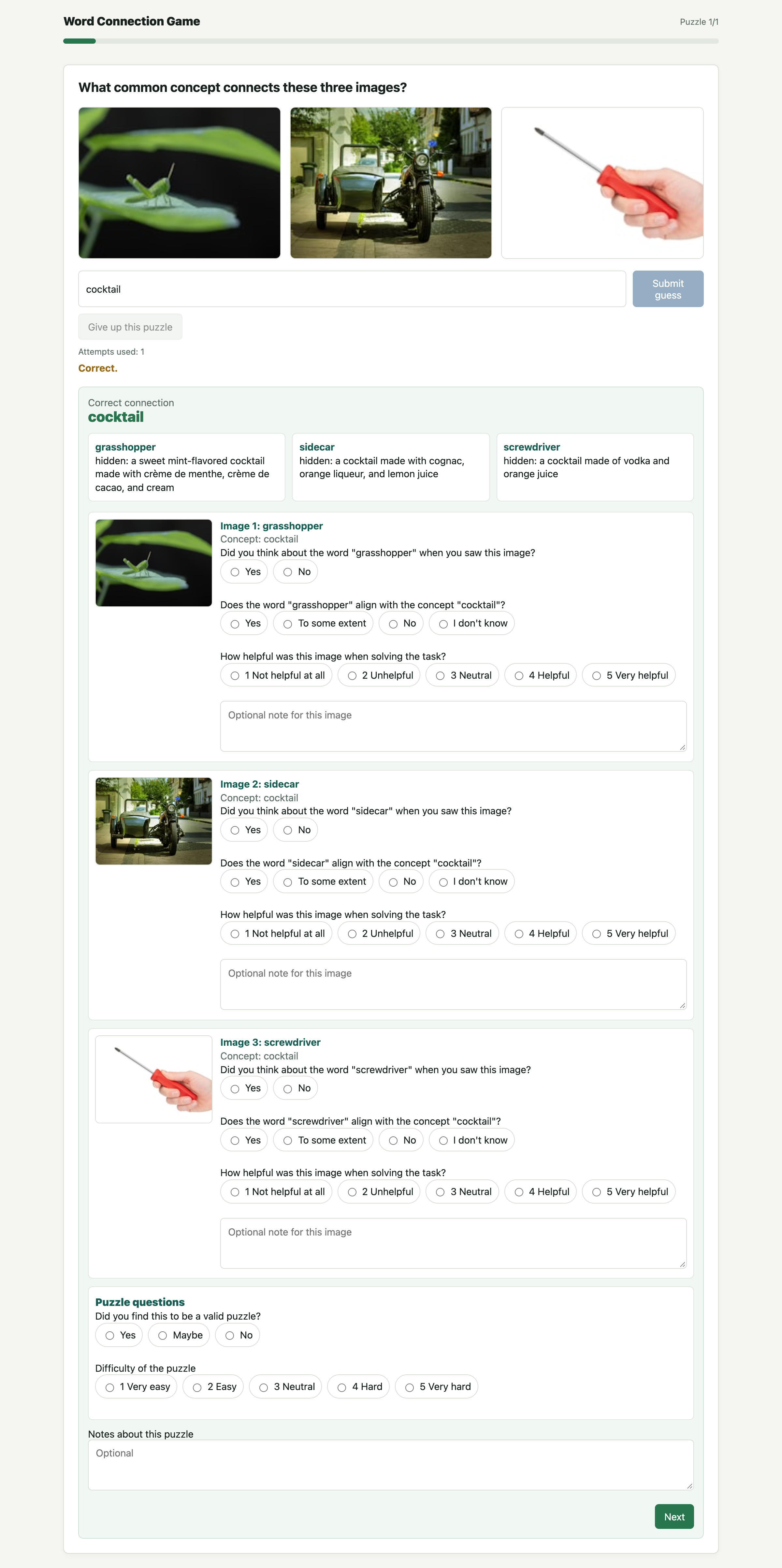}
    \caption{User interface used to conduct the human study. The image presents the questionnaire presented after the participant's attempt to solve the puzzle. The question "\textit{How helpful is this image when solving the task?}" only appears if the participant solved the puzzle.}
    \label{fig:ui_after}
\end{figure*}

Figure~\ref{fig:ui_after} shows the human study interface after revealing the intended shared concept.

\section{Reproducibility Statement}

\subsection{Model Size}

We ran six open-weight models using vLLM and NVIDIA A100 GPUs with 40GB memory each: \texttt{Molmo2-8B}, \texttt{Qwen3.5-9B}, \texttt{Qwen3.8-27B}, \texttt{Gemma-4-12B-it}, \texttt{InternVL3.5-8B}, and \texttt{Llama-4-Scout}.
\texttt{Qwen3.8-27B} and \texttt{Llama-4-Scout} used two GPUs with tensor parallelism of two. The other four models used
one GPU each. \texttt{Llama-4-Scout} has 109B total and 17B
active parameters, and we used its W4A16 checkpoint.

\subsection{Experimental Setup And Hyperparameters}
We used pretrained checkpoints without fine-tuning. Our local launcher defaults were bfloat16 computation, a 32,768-token context limit, a maximum of 8,192 batched tokens, one concurrent sequence, and 0.90 GPU memory utilization. Some launch commands used the context limit of 16,384 and the memory utilization limit of 0.95.

We used a 512-token output limit for Qwen and a 1,024-token limit for Gemma, InternVL, and Llama Scout. Molmo's output limit was left to the server default. Thinking was disabled for Qwen and Gemma. The evaluator did not override temperature, top-$p$, or top-$k$, and the unseeded repeated runs did not supply a request seed.

\end{document}